\def\year{2018}\relax
\documentclass[letterpaper]{article} %DO NOT CHANGE THIS

\usepackage{aaai18}  %Required
\usepackage{times}  %Required
\usepackage{helvet}  %Required
\usepackage{courier}  %Required
\usepackage{url}  %Required
\usepackage{graphicx}  %Required
\usepackage{amsmath}
\usepackage{amssymb}
\usepackage{subfigure}
\usepackage{multirow}
\usepackage{amsthm}
\usepackage{epstopdf}
\usepackage{paralist}

\theoremstyle{plain}
\newtheorem{thm}{Theorem}
\newtheorem{lem}[thm]{Lemma}

\theoremstyle{definition}

\theoremstyle{remark}

\usepackage{algorithm}
\usepackage{algorithmic}

\begin{document}

%%%%%%%%% TITLE
\title{Coupled Deep Learning for Heterogeneous Face Recognition}

\author{Xiang Wu, Lingxiao Song, Ran He\thanks{corresponding author}, Tieniu Tan\\
Center for Research on Intelligent Perception and Computing (CRIPAC),\\
National Laboratory of Pattern Recognition (NLPR), \\
Institute of Automation, Chinese Academy of Sciences, Beijing, P. R. China, 100190\\
{\tt\small alfredxiangwu@gmail.com, \{lingxiao.song, rhe, tnt\}@nlpr.ia.ac.cn}
}

\maketitle
%\thispagestyle{empty}

%%%%%%%%% ABSTRACT
\begin{abstract}
Heterogeneous face matching is a challenge issue in face recognition due to large domain difference as well as insufficient pairwise images in different modalities during training. This paper proposes a coupled deep learning (CDL) approach for the heterogeneous face matching. CDL seeks a shared feature space in which the heterogeneous face matching problem can be approximately treated as a homogeneous face matching problem. The objective function of CDL mainly includes two parts. The first part contains a trace norm and a block-diagonal prior as relevance constraints, which not only make unpaired images from multiple modalities be clustered and correlated, but also regularize the parameters to alleviate overfitting. An approximate variational formulation is introduced to deal with the difficulties of optimizing low-rank constraint directly. The second part contains a cross modal ranking among triplet domain specific images to maximize the margin for different identities and increase data for a small amount of training samples. Besides, an alternating minimization method is employed to iteratively update the parameters of CDL. Experimental results show that CDL achieves better performance on the challenging CASIA NIR-VIS 2.0 face recognition database, the IIIT-D Sketch database, the CUHK Face Sketch (CUFS), and the CUHK Face Sketch FERET (CUFSF), which significantly outperforms state-of-the-art heterogeneous face recognition methods.
\end{abstract}

%%%%%%%%% BODY TEXT

\section{Introduction}

The performance of real-world face recognition systems suffers a lot from illumination variations, which is a traditional challenge in face recognition. The technique of near infrared (NIR) imaging provides a low-cost and effective solution to acquire high-quality images in conditions of low light or complete darkness, for which NIR imaging is widely adopted in video surveillance and user authentication applications. However, many applications require that the enrollment of face templates is based on visual (VIS) images, such as online registration and pre-enrollment using passport or ID card. Therefore, face matching between NIR and VIS images has drawn much attention in recent decades.

%\begin{figure}
%\vspace{-2mm}
%\centering
%\includegraphics[width=0.40\textwidth]{framework.pdf}
%\caption{Two-dimensional nonlinear embedding of NIR and VIS face images by coupled deep learning. Although there are large margins between different subjects, the face images of one subject are still divided into two related groups due to the appearance gap between NIR and VIS images. The low-rank relevance and cross-modal ranking are used to alleviate the semantic gap and a small number of training samples.}
%\label{fig:network}
%\vspace{-6mm}
%\end{figure}

One major challenge in matching faces from VIS-NIR heterogeneous conditions is that images of the same subject may differ in appearance due to the changes between VIS images and NIR images. This difference introduces high intra-class variations, which generally make a direct comparison between samples have poor matching accuracy. Some research efforts have been devoted to addressing this challenge~\cite{DBLP:conf/cvpr/LiYLL13,lin_tang2006,socolinsky2002TIR_analysis,yi2007face_NIS-VIR}, and mainly focus on transforming data from different modalities onto a common comparable space. Since the changes of face appearance are often influenced by many factors (e.g. identities, illuminations and expressions) and noise information in different modalities show diverse distributions, reducing the intra-class variations of heterogeneous face data is more complex than that of homogeneous face data.

%A lot of research efforts have been devoted to solving this cross-model face matching problem~\cite{DBLP:conf/cvpr/LiYLL13,lin_tang2006,socolinsky2002TIR_analysis,yi2007face_NIS-VIR}, and they mainly focus on transforming data from different modalities onto a common comparable space. In practice face appearance is influenced by many factors (e.g. identities, illuminations and expressions). And in the cross-modal case, noise information (e.g. illuminations, expressions and resolutions) in different modality show diverse distribution. Thus, preserving the identity relevance between different modalities is more complex than general face recognition, which has not been studied well.

%Besides, benefitting from the prosperity of social networks, large-scale VIS face data are easily obtained from the Internet, which is exactly a key factor of the success of deep learning based VIS face recognition systems~\cite{sun2014deep}. As the scale of training data dramatically increases, deep learning networks become deeper and larger, resulting in a larger parameter space. However, it is hard and expensive to collect corresponding heterogeneous NIR image data. A small scale of pairwise VIS-NIR limits the application of deep learning methods to the VIS-NIR matching problem. Consequently, how to utilize these massive unpaired VIS data effectively to boost the research of heterogeneous face recognition is still an ongoing issue.

Motivated by these observations, this paper proposes a deep learning framework, named coupled deep learning (CDL), to address the VIS-NIR heterogeneous matching problem. It transfers the deep representation learnt on a large-scale VIS dataset~\cite{wulight} to adapt to NIR domain by introducing a novel VIS-NIR objective function for convolution neural networks. It seeks a deep feature space in which the heterogeneous face matching problem can be approximately treated as a homogeneous face matching problem. Inspired by the relevance constraints in cross-modal learning~\cite{DBLP:conf/iccv/WangHWWT13}, we introduce a trace norm and a block-diagonal prior to the softmax loss. These relevance constraints can enhance the relevance of projection matrices of NIR and VIS images so that the difference between modalities is reduced and identity information is reserved, especially for the small number of training samples. On the account of making use of the massive VIS face images, a cross modal ranking loss defined on a set of NIR-VIS triplets is introduced. Based on the alternative formulation for the trace norm, an iterative solution is developed and combined it into the convolutional neural network. Due to the limited pairwise VIS-NIR images, the parameters of shallow network layers are shared among different modalities. Experimental results on several heterogeneous face databases verify the effectiveness of each part in the CDL loss function.

%Motivated by these observations, a Coupled Lightened Convolution Neural Network (CL-CNN) framework is proposed in this paper to learn modality-invariant representation of face images as shown in Fig.~\ref{fig:network}. Firstly, to alleviate the difference between modalities, we introduced lightened CNN\cite{DBLP:journals/corr/WuHS15} to map both NIR and VIS data to a common latent space under the constraint of low-rank, which helps to reserve identity information and remove noise information. Secondly, on the account of making use of the massive unpaired VIS face images, a cross modal ranking loss defined on a set of triplets is introduced, and the parameters of shallow network layers are shared among different modalities. Finally, based on the alternative formulation for the trace norm, an iterative solution is developed and combined it into the convolutional neural network.

The main contributions are summarized as follows:
\begin{itemize}\setlength{\itemsep}{1pt}
  \item Two novel relevance constraints on fully connected layer, as regularizers, are used not only to enforce the relevance between different modalities, but also to constrain the parameter space, which can alleviate overfitting especially on a small amount of unpaired heterogenous samples.
  \item Due to the difficulty to directly optimize the high-order trace norm, we introduce an approximate variational formulation of the trace norm and present an alternating algorithm to efficiently optimize it in an end-to-end CNN.
  \item A cross-modal ranking sampling method defined on a set of cross-modal triplets is applied to maximize the margin between different subjects. Besides, it is effective to enlarge the training data and to utilize the information between a limited number of heterogenous samples.
  \item Extensive experimental evaluations on the challenging CASIA NIR-VIS 2.0 face database and three viewed sketch-photo databases show that the proposed method improves the state-of-the-art performance on heterogeneous face recognition.
\end{itemize}

The remainder of the paper is organized as follows. In Section 2, we review related work on the heterogeneous face recognition problem. Section 3 describes our proposed coupled deep learning framework for heterogeneous face recognition. In Section 4, we report experimental results on the different heterogeneous face recognition datasets. Finally, we conclude the paper in Section 5.

\section{Related Work}
\subsection{Heterogeneous Face Recognition}
The task of heterogeneous face recognition is to match face images that come from different modalities. %, such as visual versus near infrared (VIS-NIR) face recognition~\cite{yi2009partial_NIS-VIR,yi2007face_NIS-VIR}, visual versus thermal infrared face recognition~\cite{socolinsky2002TIR_analysis,socolinsky2003face_TIR-VIS}, face photo versus face sketch~\cite{tang2004face_sketch,DBLP:journals/pami/WangT09}.
%According to the way to alleviate the appearance difference between heterogeneous data,
Existing heterogeneous face recognition methods can be roughly divided into three categories~\cite{zhu2014matching}: (i) data synthesis based; (ii) latent subspace learning based; (iii) modal-invariant feature learning based methods.

\textbf{Data synthesis} based methods aim to project the data of one modality into the space of another modality by data synthesis. Thus the similarity relationship of heterogeneous data from different domain can be measured. Liu et al.~\cite{liu2005nonlinear} propose a local geometry preserving based nonlinear method to generate pseudo-sketch from face photo. Wang \emph{et al.}~\cite{DBLP:journals/pami/WangT09} present a cross-spectrum face mapping method to transform NIR and VIS data to another type, and thereby perform face matching. Wang and Tang~\cite{DBLP:journals/pami/WangT09} use a multi-scale Markov Random Fields (MRF) model to synthesize sketch drawing from given face photo and vice versa. Lei \emph{et al.}~\cite{lei2008CCA_mapping} propose a canonical correlation analysis (CCA) based multi-variant mapping algorithm to reconstruct 3D model from a single 2D NIR image. %Other works~\cite{huang2013coupled,DBLP:conf/cvpr/Juefei-XuPS15,wang2012sketch_synthesis} resort to coupled or joint dictionary learning to reconstruct face images, and then perform face matching. The main limitation of data synthesis based methods is that large amount of pairwise heterogeneous data is needed in training process.
%Gao \emph{et al.}~\cite{gao2008face_sketch} further study face sketch synthesis using embedded hidden Markov model (E-HMM) and selective ensemble strategy.
%applied face analogy to transform a face image from one modality to another.

\textbf{Latent subspace learning} based approaches project both modalities to a common latent space, in which the relevance of data from different modalities can be measured. Dimension reduction techniques such as Principal Component Analysis (PCA), Canonical Correlation Analysis (CCA) and Partial Least Squares (PLS) are often used. Lin and Tang~\cite{lin_tang2006} propose a method called Common Discriminant Feature Extraction (CDFE) method transform data to a common feature space. CDFE takes both inter-modality discriminant information and intra-modality local consistency into consideration.
%Yi \emph{et al.}\cite{DBLP:conf/icb/YiLLSL09} apply CCA to learn the best correlation regression between NIR and VIS faces from NIR-VIS face pairs. Lei \emph{et al.}~\cite{lei2012coupled} present a coupled discriminant analysis method that involves the locality information in kernel space. Huang \emph{et al.}~\cite{huang2013regularized} propose a regularized discriminative spectral regression (DSR) method to map heterogeneous data. Recently, some works attempt to address the cross-modal matching problem by deep learning.
Hou \emph{et al.}~\cite{hou2014domain} propose a domain adaptive self-taught learning approach to derive a common subspace. Yi \emph{et al.}~\cite{DBLP:conf/fgr/YiLL15} suggests to use Restricted Boltzmann Machines (RBMs) to learn a shared representation between different modalities, and then apply PCA to remove the redundancy and heterogeneity. The state-of-the-art NIR-VIS result on the CASIA NIR-VIS 2.0 face database is now obtained by \cite{DBLP:conf/fgr/YiLL15}.
%
% State-of-the-art NIR-VIS results are often obtained by removing some principal subspace components Yi et al.~\cite{yidong2015shared}. Recently, Wang et al.~\cite{wangkaiye2013iccv} took feature selection into consideration during common subspace learning.

\textbf{Modality-invariant feature learning} based methods seek modality-invariant features that are only related to face identity. Existing algorithms in this category are almost based on handcrafted local features, such as local binary patterns (LBP), Histograms of Oriented Gradients (HOG) and Difference of Gaussian (DoG). Liao \emph{et al.}~\cite{liao2009heterogeneous} use DoG filtering as preprocessing for illumination normalization, and then employ Multi-block LBP (MB-LBP) to encode NIR as well as VIS images. Klare \emph{et al.}~\cite{klare2010heterogeneous} further combine HoG features to LBP descriptors, and utilize sparse representation to improve recognition accuracy. HoG feature and its variants are applied to NIR-VIS face matching in \cite{dhamecha2014effectiveness}. Goswami \emph{et al.}~\cite{goswami2011evaluation} incorporate a series of preprocessing methods to perform normalization, then combine a Local Binary Pattern Histogram (LBPH) representation with LDA to extract invariant features. However, most of above features are designed empirically. Zhu \emph{et al.}~\cite{zhu2014matching} involve Log-DoG filtering, local encoding and uniform feature normalization together to reduce the appearance difference between VIS and NIR images. %They also provide theoretical analysis based on an illumination reflectance model.
  %~\cite{liao2009heterogeneous,klare2010heterogeneous,goswami2011evaluation} In addition, Huang et al.~\cite{huang_lu_tan2012learning} utilize sparse representation to learn modality-invariant features.
 %Goswami \emph{et al.}~\cite{goswami2011evaluation} utilize local binary pattern histogram representation in tandem with LDA for cross spectral matching.

Although some heterogeneous face recognition algorithms have obtained good results, the performance of heterogeneous face matching tasks is still far below than that of VIS face matching, which benefits fully from the development of deep learning models. In the last decades, numerous deep learning based face recognition algorithms have been proposed~\cite{liu2016viplfacenet,schroff2015facenet,parkhi2015deep,wen2016discriminative,wulight}, and many of them have more than $98\%$ verification accuracy on the challenging LFW database. However, there are few deep learning methods for heterogeneous face recognition due to the lack of training samples.

\subsection{Cross-view Classification}

Multi-view learning involves relating information from multiple sources or views. Ngiam \emph{et al.} \cite{DBLP:conf/icml/NgiamKKNLN11} considered a shared representation learning setting from multiple modalities with RBM. Wang \emph{et al.} \cite{DBLP:conf/icml/WangALB15} proposed several DNN-based methods with linear or kernel CCA in unsupervised multi-view feature learning. However, the second view is used as the prior and is not available at test time. Yan \cite{DBLP:conf/cvpr/YanM15} address the problem of matching images and captions with deep canonical correlation analysis. They employ convolution networks to obtain representations and then seek pairs of linear projections by maximizing the correlation of the two views according to CCA. Kan \cite{DBLP:conf/cvpr/KanSC16} also attempts to seek for a discriminant and view-invariant representation by analyzing the within-class and between-class scatter. Besides, Maximum Mean Discrepancy (MMD) \cite{MMD_gretton2012kernel} is widely used to learn domain-invariant representations by minimizing the domain discrepancy, but it only considers the first momnent (mean) of the multiple modal distributions. Central Moment Discrepancy \cite{DBLP:journals/corr/ZellingerGLNS17} introduce high-order moments as a metric in the field of domain-invariant representation learning.

Since the existing cross-view methods are unsupervised deep non-linear methods or generic cross-view classification methods. While heterogenous face recognition is zero-shot learning which the identities in testing dataset is exclusive from the identities in training dataset and it is also treated as fine-grained classification because samples are similar even in different modalities. Due to these properties, generic or unsupervised cross-view methods cannot find a generalized embedding space to sperate each identity from different modalities. Exploring an explicit cross-modal methods for heterogeneous face matching problem via deep learning deserves to be further studied.

\section{Method}

In this section, we first introduce the two major parts in the CDL objective function, and then we detail the whole network architecture of CDL.

\subsection{Relevance Constraints}

%In cross-modal learning \cite{DBLP:conf/ijcai/LiangHST16,DBLP:conf/iccv/WangHWWT13}, a key issue is to measure the correlation between multi-modal data or their corresponding projection matrices. The projection matrices between data representation and a supervised signal are often assumed to be correlated or share some common structures.
In VIS-NIR face recognition, the projection matrices are often learnt by subspace learning methods. Particularly, PCA is used as a post-processing step to improve matching accuracy. Inspired by these observations, we introduce two relevance constraints to the softmax loss of CNN.

For the first one, given $\text{I}_V$ and $\text{I}_N$ for VIS and NIR images respectively, we can denote the CNN feature extraction process as $X_i=\text{Conv}(\text{I}_i, \Theta_i), i\in \{N, V\}$, where Conv() is defined by the convolution neural network, $\Theta_i$ denotes the CNN parameters and $X_i\in \mathrm{R}^m$ is the feature representation. In heterogeneous face recognition, we assume that there are some common low-level features between NIR and VIS images. Hence, it is $\Theta_N=\Theta_V=\Theta$ and the representation can be shown as $X_i=\text{Conv}(\text{I}_i, \Theta)$ and we also denote that $W_i, i\in\{N, V\}$ is the parameters of last fully connected layer, treated as a classifier, for NIR and VIS samples, respectively.

%As the softmax loss is widely used as the supervised signal for CNN, we can define a generic loss function in the following form,
%\begin{equation}
%\begin{split}
%\mathcal{J}(X_i, W_i) &= \sum_{i\in\{N,V\}}\text{softmax}(X_i, W_i) \\
%            &=-\sum_{i\in\{N,V\}}(\sum_{j=1}^N1\{y_{ij}=c\}\text{log}{\hat{p}_{ij}})\\
%\end{split}
%\end{equation}
%where $W_i, i\in\{N, V\}$ is the parameters for NIR and VIS samples, respectively, $c$ is the class label for each sample and $\hat{p}_{ij}$ is the predicted probability. Besides, we define $1\{\cdot\}$ is the indicator function so that $1\{\text{a true statement}\}=1$ and $1\{\text{a false statement}\}=0$.

Since the core of cross-modal matching is searching the common feature space to reserve the discriminative information, we impose a trace norm to the softmax loss
\begin{equation}
\label{relevance_constraint}
\mathcal{R}_1 = \|\left[W_N \ \ W_V\right]\|_*
\end{equation}
where $\|\cdot\|_*$ is the trace norm, which is used to enforce the relevance of the projected common label space. The trace norm in Eq. (\ref{relevance_constraint}) can also reduce the parameter space of CNN for a small-scale dataset.

Eq.~(\ref{relevance_constraint}) is difficult to directly optimize for the presence of the trace norm. Here, we introduce an iterative minimization solution for the trace norm\cite{DBLP:conf/nips/GraveOB11} based on its variational formulation.

\begin{lem}
Let $\mathrm{M} \in \mathrm{R}^{m\times n}$. The trace norm of $\mathrm{M}$ is shown as:
\begin{equation}
\|\mathrm{M}\|_*=\sum_{i=1}^{\min(m,n)}\sigma_i
\end{equation}
where $\sigma_i$ denotes the $i$-th singular value of $\mathrm{M}$. The formulation can be equal to:
\begin{equation}
\|\mathrm{M}\|_*=\frac{1}{2}\inf_{\Gamma\geq 0}\mathrm{tr}(\mathrm{M}^T\Gamma^{-1}\mathrm{M}) + \mathrm{tr}(\Gamma)
\end{equation}
and the semi-definite matrix $\Gamma$ is attained for $\Gamma=(\mathrm{MM}^T)^{\frac{1}{2}}$.
\end{lem}

According to this lemma, the trace norm constraint can be reformulate as
\begin{equation}
\mathcal{R}_{1} = \frac{1}{2}\lambda\text{tr}(\left[W_N \ \ W_V\right]^T\Gamma^{-1}\left[W_N \ \ W_V\right]) + \frac{1}{2}\lambda\text{tr}(\Gamma)\\
\label{reformulation_relevance_constraint}
\end{equation}
%In terms of the trace properties, the second term in Eq.~(\ref{reformulation_relevance_constraint}) can be represented as:
%\begin{equation}
%\begin{split}
%&\text{tr}(\left[W_N \ \ W_V\right]^T\Gamma^{-1}\left[W_N \ \ W_V\right])\ \\
%&= \text{tr}(W_N^T\Gamma^{-1} W_N) + \text{tr}(W_V^T\Gamma^{-1} W_V)
%\end{split}
%\end{equation}
and according to \cite{DBLP:conf/nips/GraveOB11}, the infimum over $\Gamma$ can be computed as
\begin{equation}
\Gamma=(W_NW_N^T + W_VW_V^T+\mu\mathrm{I})^{\frac{1}{2}}
\label{update_gamma}
\end{equation}

To minimize Eq.~(\ref{reformulation_relevance_constraint}) in CNN, we develop an alternating minimization method by updating the parameters $\Theta, W_i$ and $\Gamma$. For the convolution parameter $\Theta$, we follow the conventional back-propagation method to update it. For the matrix $\Gamma$, we can update it via Eq.~(\ref{update_gamma}). For the projected matrix $W_V$ and $W_N$, the gradients contain two components which can be expressed as
\begin{equation}
\frac{\partial \mathcal{R}_1}{\partial W_N}=  W_N(\Gamma^{-1}+(\Gamma^{-1})^T)
\label{gradient_WN}
\end{equation}
\begin{equation}
\frac{\partial \mathcal{R}_1}{\partial W_V}=  W_V(\Gamma^{-1}+(\Gamma^{-1})^T)
\label{gradient_WV}
\end{equation}

To obtain  $\Gamma^{-1}$ for updating $W_N$ and $W_V$, we can compute $U\text{Diag}(\gamma_k)V^T$ by the singular value decomposition of $W_NW_N^T + W_VW_V^T$. Hence, the inverse of matrix $\Gamma$ takes the following form
\begin{equation}
\Gamma^{-1}=V\text{Diag}(\frac{1}{\sqrt{(\gamma_k+\mu)}})U^T
\end{equation}

For the second relevance constraint, we define  block-diagonal prior constraint for two fully connected matrices $W_V$ and $W_N$, respectively. The block-diagonal prior can efficiently regularize the parameters to alleviate the overfitting during training on the small number of samples. We can define the orthogonal regularizer as:
\begin{equation}\label{orth_const}
\mathcal{R}_2 = \frac{1}{2}\left[\|W_N^TW_N-I\|_F^2 + \|W_V^TW_V-I\|_F^2\right]
\end{equation}
where the $\|\cdot \|_F^2$ is Frobenius norm. Therefore, the gradients can be written as:
\begin{equation}\label{gradient_F_2}
\begin{split}
\frac{\partial \mathcal{R}_2}{\partial W_N} &=  W_N(W_N^TW_N)-W_N\\
\frac{\partial \mathcal{R}_2}{\partial W_V} &=  W_V(W_V^TW_V)-W_V\\
\end{split}
\end{equation}

Therefor, the softmax loss with relevance constraints can be defined as
\begin{equation}
\mathcal{J}_{\text{relevance}} = \text{softmax}(X, W_V, W_N, \Theta) + \alpha_1\mathcal{R}_1+\alpha_2\mathcal{R}_2
\end{equation}

%Then the parameters can be updated with the learning rate $\alpha$ as follows.
%\begin{align}
%\Theta^{(t+1)} &= \Theta^{(t)} -\alpha \frac{\partial \mathcal{J_\text{relevance}}}{\partial \Theta^{(t)}}\\
%W_N^{(t+1)} &= W_N^{(t)} -\alpha \frac{\partial \mathcal{J_\text{relevance}}}{\partial W_N^{(t)}}\\
%W_V^{(t+1)} &= W_V^{(t)} -\alpha \frac{\partial \mathcal{J_\text{relevance}}}{\partial W_V^{(t)}}
%\end{align}

%\begin{algorithm}[tb]
%\caption{Training the CNN with relevance constraint.}
%\label{algorithm1}
%\begin{algorithmic}[1]
%\REQUIRE
%Training set: Training set $I_i$, learning rate $\alpha$ and the trade-off parameter $\lambda$.
%\ENSURE
%The CNN parameters $\Theta$, the projected matrices $W_N, W_V$ and the covariance matrices $\Gamma$.
%\STATE Initialize parameters $\Theta, W_N, W_V$;
%\FOR {$t=1,\dots, T$}
%\STATE Fix $\Theta, W_N, W_V$;
%\STATE \quad Update $\Gamma$ according to Eq.~(\ref{update_gamma});
%\STATE Fix $\Theta, \Gamma$
%\STATE \quad Update $W_N, W_V$ according to Eq.~(\ref{gradient_WN}) and Eq.~(\ref{gradient_WV});
%\STATE Fix $W_N, W_V, \Gamma$
%\STATE \quad Update $\Theta$ according to the back-propagation method;
%\ENDFOR;
%\STATE \textbf{Return} $W_N, W_V$ and $\Theta$;
%\end{algorithmic}
%\end{algorithm}

\subsection{Cross Modal Ranking}

Considering the cross modal matching between NIR and VIS face images, encouraged by \cite{schroff2015facenet}, we formulate a cross modal ranking term to preserve the intra-personal similarity in different domains. When the cross modal triplet ranking regularization is employed, both NIR and VIS face images of one identity are potentially encouraged to be projected onto the same point in the modal invariant embedding subspace.

Moreover, the deep convolution neural network usually requires a large amount of training data, however, there are not enough labeled NIR and VIS images for training by the softmax loss. The cross modal triplet ranking can enlarge the number of training data.

Given $x_i\in X=\{X_N, X_V\}$, where $x_{N_i}\in X_N$ and $x_{V_i}\in X_V$ for NIR and VIS features respectively, we can sample a set of triplet tuples depending on the labels $T_i=(x_i^a, x_i^p, x_i^n)$, where $x_i^a$ and $x_i^p$ are images of the same label, while $x_i^p$ and $x_i^n$ are with different labels. For a triplet tuple, the ranking loss is defined as
\begin{equation}
\mathcal{J}_{\text{ranking}}=\sum_{i=0}^N\max(0, m + \|x_i^a-x_i^p\|^2 - \|x_i^a - x_i^n\|^2)
\label{triplet_constraint}
\end{equation}
where the constant parameter $m$ defines a margin between positive and negative pairs.

It is crucial to select the triplets, which are active and can contribute to improve model performance. Since the triplet ranking is used to reduce the gap between different modalities, random selection of triplet tuples is not applicable.
Therefore, we propose a strategy for cross modal triplets selection.
The anchor sample $x^a_i$ and the negative one $x^n_i$ is selected from the same modality and the modality of positive sample $x_i^p$ is different from anchor and negative ones. For exmaple, if the anchor sample is selected from near infrared set $X_N$, the positive one is from the visible light set $X_V$ and the negative one is from $X_N$.

Moreover, a hard triplet tuple may be caused by mislabeling or poor image quality, so it might lead to poor training. On contrast, a simple triplet is easy to satisfy the triplet constraint in Eq.~(\ref{triplet_constraint}) and less contributes to the training so that the network converges slowly. Therefore, the semi-hard triplet selection method is adopted to further improve the cross modal matching performance.

From the above, the cross modal triplet selection is based on the constraints as follows:
\begin{equation}
\left\{\begin{array}{c}
\|x_i^a-x_i^p\|^2 + m > \|x_i^a - x_i^n\|^2\\
\\
\|x_i^a-x_i^p\|^2 < \|x_i^a - x_i^n\|^2\\
\\
{x_i^a, x_i^n} \in X_N(X_V), ~~{x_i^p} \in X_V(X_N)
\end{array}
\right.
\label{tripet_selection}
\end{equation}
where the constant parameter $m$ is the margin in Eq.~(\ref{triplet_constraint}). Under these constraints shown in Eq.~(\ref{tripet_selection}), the network focuses more attention on the individual distinction to eliminate cross modal variance.

\subsection{Network Architecture}

We employ the light CNN \cite{wulight} as our basic network. The network includes nine convolution layers, four max-pooling layers and one fully-connected layer. Softmax is used as the loss function. To address the small-scale data for NIR-VIS training, we firstly train a CNN on the large visible light face dataset and then fine-tune the NIR-VIS one on the pre-trained visible light face model. The basic VIS network and initial values of $\Theta$ are trained on the MS-Celeb-1M dataset\cite{DBLP:journals/corr/GuoZHHG16} which contains 100K identities totally about 8.5M images. %The training face images are normalized and cropped to $144\times 144$ according to five facial landmarks. To enrich the training data, the input images are randomly cropped into $128\times 128$.

Based on the basic network, we develop a coupled deep learning (\textbf{CDL}) framework for NIR-VIS face recognition. We combine the softmax term with relevance constraints and cross modal ranking term as the supervised signal:
\begin{equation}
\mathcal{J} = \lambda_1\mathcal{J}_{\text{relevance}} + \lambda_2\mathcal{J}_{\text{ranking}}
\label{objective_function}
\end{equation}
where $\lambda_1, \lambda_2$ is the trade-off between two loss terms. The softmax with relevance constraint term aims to enforce the correlation across different modalities for each identity. And the cross modal ranking can not only enforce the intrinsic properties of the same identities from different domain, but also enlarge the inter-personal similarity in the same domain. The gradients of all the parameters can be computed by
\begin{equation}
\label{grad_theta}
\frac{\partial\mathcal{J}}{\partial\Theta}=\lambda_1\frac{\partial \mathcal{J_\text{relevance}}}{\partial \Theta}+\lambda_2\frac{\partial \mathcal{J_\text{ranking}}}{\partial \Theta}
\end{equation}
\begin{equation}
\label{grad_wn}
\frac{\partial \mathcal{J}}{\partial W_N}=\lambda_1\frac{\partial \mathcal{J_\text{relevance}}}{\partial W_N}
\end{equation}
\begin{equation}
\label{grad_wv}
\frac{\partial \mathcal{J}}{\partial W_V}=\lambda_1\frac{\partial \mathcal{J_\text{relevance}}}{\partial W_V}
\end{equation}

According to Eq.~(\ref{objective_function}), the goal of CDL is to learn the parameters $\Theta$ for feature extraction, while $W_N, W_V$ and $\Gamma$ are only parameters introduced to propagate signals during training. All the parameters are updated by stochastic gradient descent and an alternating minimization method is proposed to update $W_N, W_V$ and $\Gamma$. The details of CDL training is shown in Algorithm \ref{algorithm2}.

\begin{algorithm}[tb]
\caption{Coupled Deep Learning (CDL) Training.}
\label{algorithm2}
\begin{algorithmic}[1]
\REQUIRE
Training set: NIR images $I_N$, VIS images $I_V$, the learning rate $\alpha$, the ranking threshold $m$ and the trade-off parameters $\lambda, \lambda_1, \lambda_2$.
\ENSURE
The CNN parameters $\Theta$.
\STATE Initialize parameters $\Theta, W_N, W_V$ by pre-trained VIS model;
\FOR {$t=1,\dots, T$}
\STATE Forward propagation to obtain $\mathcal{J}_{\text{relevance}}$ and $\mathcal{J}_{\text{ranking}}$;
\STATE Compute gradients according to Eq.~(\ref{grad_theta}), Eq.~(\ref{grad_wn}) and Eq.~(\ref{grad_wv});
\STATE Fix $\Theta, W_N, W_V$;
\STATE \quad Update $\Gamma$ by Eq.~(\ref{update_gamma});
\STATE Backward propagation for $\Theta, W_N, W_V$;
\STATE ~Fix $W_N, W_V, \Gamma$
\STATE ~\quad Update $\Theta$ by Eq.~(\ref{grad_theta});
\STATE ~Fix $\Theta, \Gamma$
\STATE ~\quad Update $W_N, W_V$ by Eq.~(\ref{grad_wn}) and Eq.~(\ref{grad_wv});
\ENDFOR;
\STATE \textbf{Return} $\Theta$;
\end{algorithmic}
\end{algorithm}

\begin{figure*}[tbp]
\centering
\subfigure[intra-class variance]{\includegraphics[width=0.31\textwidth]{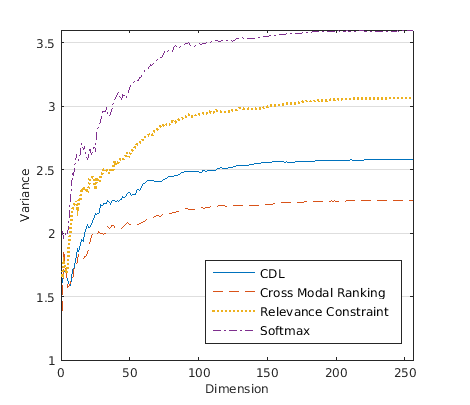}\label{intra_class}}
\subfigure[inter-class variance]{\includegraphics[width=0.31\textwidth]{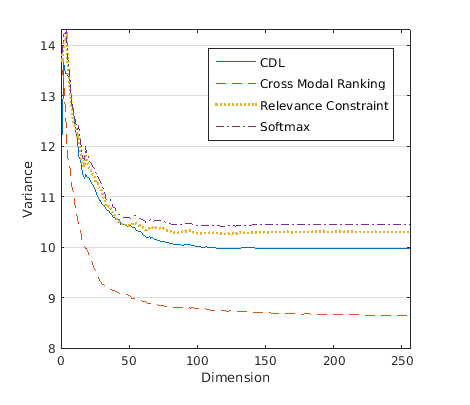}\label{inter_class}}
\subfigure[The correlation matrix of last fully connected layer]{\includegraphics[width=0.32\textwidth]{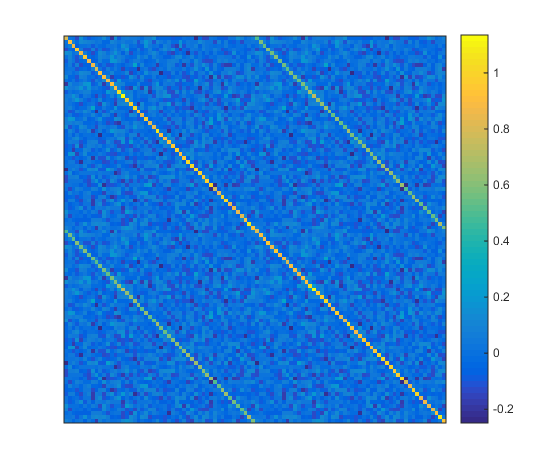}\label{relevance}}
\caption{The performance analysis of CDL. (a) and (b) shows the variations of intra-class and inter-class distance respectively. The x-axis indicates the dimension of features by PCA and the y-axis indicates intra-class and inter-class variances respectively. (c) shows the correlation matrix of $\left[W_N \ \ W_V\right]$, the lighter color indicates high correlations.}
\label{fig:inter_intra_class}
\vspace{-4mm}
\end{figure*}

\section{Experiments}

In this section, we evaluate the proposed CDL on the most challenge CASIA NIR-VIS 2.0 face database \cite{DBLP:conf/cvpr/LiYLL13}, CUHK Face Sketch (CUFS) \cite{DBLP:journals/pami/WangT09}, CUHK Face Sketch FERET (CUFSF) \cite{DBLP:conf/cvpr/ZhangWT11} and IIIT-D Sketch Database \cite{bhatt2012memetic}. First, we introduce the database and protocols. Second, the details of training methodology are presented. Then we perform the algorithmic analysis and compare our method with other state-of-the-art heterogeneous face recognition methods.

\subsection{Dataset and Protocol}
The CASIA NIR-VIS 2.0 face database is widely used to evaluate heterogeneous face recognition algorithms. Its challenge contains large variations of the same identity, expression, pose and distance. The database collects 725 subjects, each with 1-22 VIS and 5-50 NIR images and all the images are randomly gathered, therefore, there are not one-to-one correlations between NIR and VIS images.
There are two views of evaluation protocols for the database. View 1 is used for super-parameters adjustment, and View 2 is used for training and testing.
For a fair comparison with other methods, we choose the standard protocol in View 2. There are 10-fold experiments in View 2. Each fold contains training and testing lists. Nearly equal numbers of identities are included in the training and testing sets. For each fold, there are about 6,100 NIR images and 2,500 VIS images from about 360 identities. These subjects are exclusive from the 358 identities in the testing set. For the testing of each fold, the gallery contains 358 identities and each identity has one VIS image. The probe has over 6,000 NIR images from the same 358 identity. All the NIR images in the probe set compare against the gallery set and the evaluations are performed on verification rate(VR)@false acceptance rate(FAR) and rank-1 accuracy.

The IIIT-D Sketch database, the CUHK Face Sketch (CUFS), and the CUHK Face Sketch FERET (CUFSF) are all the viewed sketch-photo face database. the IIIT-D Sketch database comprises a total of 238 sketch-photo image pairs. The sketches are drawn by a professional sketch artist for digital images collected from different sources which consist of 67 pairs from the FG-NET aging database, 99 pairs from Labeled Faces in Wild (LFW) database, and 72 pairs from the IIIT-D student \& staff database. The CUFS contains 306 persons totally 712 face images for training and 300 persons totally 600 face images for testing. Besides, on CUFSF, there are 500 persons for training and 694 persons for testing. Due to the few number of images in CUFS and the IIIT-D Sketch database, we use CUFSF as the training dataset and then the rank-1 accuracy for probe-gallery face identification testing is used for them. For the CUFSF, we follow its training and testing protocols to evaluate the proposed CDL methods.

\subsection{Training Methodology}

First, we train the basic light CNN on the MS-Celeb-1M dataset. All the training face images are converted to gray-scale and normalized to $144\times 144$ according to five facial landmarks \cite{wulight}. Then we crop an input image into $128\times128$ randomly and mirror it to enrich the number of training data. The momentum is set to 0.9 and the weight decay is set to 5e-4. Moreover, the drop ratio for the fully connected layer is set to 0.7. The learning rate is set to 1e-3 initially and reduced to 5e-5 gradually and we initialize the convolution parameters by Xavier and the fully-connected layers by Gaussian, respectively. The model is trained on TITAN X for two weeks and the performance on LFW obtains \textbf{98.80\%}.

Based on the basic VIS network, we implement CDL on the heterogeneous face recognition database. CDL is initialized by the pre-trained basic model. For network training, the input images are also normalized to $144\times 144$ and randomly cropped to $128\times 128$, which is the same as the basic network. The batch size is set to 128 and the learning rate is decreased from 1e-4 to 1e-6 gradually for around 200,000 iterations. The trade-off parameter $\lambda_1$ for softmax term is set to 1 and $\lambda_2$ for cross modal ranking term is increased from 0 to 1 gradually. Moreover, the constant $\lambda$ for relevance constraint in softmax is set to 0.001.

\subsection{Algorithmic Analysis}

\begin{table}[tbp]
\centering
\caption{The verification rate(VR)@false accept rate(FAR) ($\pm$ standard variation) of CDL with different supervised terms for the CASIA NIR-VIS 2.0 database.}
\begin{tabular}{|c|c|c|}
\hline
Supervised term   &  FAR=0.1\% & FAR=0.01\% \\
\hline
\hline
Basic model & 85.31$\pm$0.95 &63.52$\pm$3.14\\
\hline
Softmax  & 93.66$\pm$1.74 &80.76$\pm$5.29\\
\hline
Trace Norm & 94.57$\pm$1.51 &82.96$\pm$4.80\\
\hline
Block-diagonal Prior  & 95.07$\pm$0.81 &83.62$\pm$3.70\\
\hline
Cross Modal Ranking & 93.23$\pm$1.05 & 78.93$\pm$4.11\\
%\hline
%Softmax + Central Moment Discrepancy &95.93$\pm$0.53 & 94.07$\pm$0.53 &78.26$\pm$2.68\\
%\hline
%Softmax + Cross Modal Ranking  &98.29$\pm$0.42 &97.11$\pm$0.33 &91.03$\pm$0.96\\
%\hline
%CDL (non-shared convolution) & 98.36\%$\pm$0.31\% &98.02\%$\pm$0.18\% &92.07\%$\pm$0.70\% \\
\hline
CDL  & \textbf{98.32$\pm$0.05} & \textbf{93.24$\pm$0.66}\\
\hline
\end{tabular}
\label{tab:compare}
\end{table}

The aim of the proposed CDL is to remove the gap between different spectral domains and find the intrinsic properties of the same identity. In this subsection, we investigate the interactions of softmax, relevance constraint and cross modal ranking on the CASIA NIR-VIS 2.0 database.

To analyze the performance of each supervised term for CDL, we introduce the concept of linear discriminant analysis (LDA) from the view of inter-class and intra-class variations. According to LDA, the intra-class variance and inter-class can be denoted as
\begin{equation}
\sigma_{\text{intra}}=\frac{1}{c}\sum_{i=1}^c\frac{1}{N_i}\sum_{x\in X_i}(x-\bar{x_i})^T(x-\bar{x_i})
\end{equation}

\begin{equation}
\sigma_{\text{inter}}=\frac{1}{N}\sum_{i=1}^c(\bar{x_i}-\bar{x})^T(\bar{x_i}-\bar{x})
\end{equation}
where $X_i$ is the set of features of the $i$-th identity, $c$ is the number of classes, $N$ and $N_i$ are the number of all the samples and ones of the $i$-th identity respectively, $\bar{x_i}$ is the corresponding mean and $\bar{x}$ is the mean of the entire dataset.

When only the softmax term is used, we find there are both diverse inter-class and intra-class variations for the learned embedding, as shown by the purple curves in Fig. \ref{fig:inter_intra_class}. Although the inter-class variations help to distinguish different identities, the large diversity of intra-class variations introduce noise, especially for the cross modal matching.
When there is only cross modal ranking as the supervised signal, as shown by the orange curve, we find that both inter-class and intra-class variance tend to decrease, since only cross modal ranking can hardly enforce a large-span feature embedding space. Besides, with the low inter-class variation, it is difficult to distinguish different identities.

The correlation matrix of $\left[W_N \ \ W_V\right]$ is shown in Fig.~\ref{relevance}. It is obviously that the elements in not only main diagonal but also bottom-left and top-right blocks have lighter color. Since the lighter color shows the higher correlations, the Fig.~\ref{relevance} presents that the relevance constraints can increase the correlations between different domains for the same identity.

As is shown in Fig.~\ref{intra_class} and Fig.~\ref{inter_class}, the yellow curve indicates the inter-class and intra-class variance by only using the relevance constraint on softmax. Obviously, it can not only decrease the intra-class variation , but also reserve the inter-class variation. It shows that the low-rank relevance constraint in Eq.~(\ref{relevance_constraint}) can enforce the correlation between different modal samples.
Furthermore, combined relevance constraint and cross modal ranking term with appropriate trade-off parameters($\lambda_1=1, \lambda_2=1$ and $\lambda=0.001$), the intra-class variance decreases in both diversity and magnitude(see by the blue curve in Fig.~\ref{intra_class} and Fig.~\ref{inter_class}). At the same time, the inter-class variation keeps nearly unchanged.
Obviously, the relevance constraint and cross modal ranking contribute to reducing the gap between different modal domains, while the softmax term can reserve the diversity of different identities instead.

%Fig.~\ref{rank_error} gives an overview of some failure cases for NIR-VIS matching. Obviously, the matching error is caused by the large pose and emotion variations. Since there are not enough NIR face images including different pose and emotions in the training dataset, CDL can not adopt these variations for NIR-VIS face matching.

\begin{table*}[htbp]
\centering
\caption{The comparison of Rank-1 accuracy ($\pm$ standard variation) and VR@FAR=0.1\% ($\pm$ standard variation) on the CASIA NIR-VIS 2.0 database.}
\begin{tabular}{|c|c|c|c|}
\hline
Method  & Rank-1 accuracy(\%) &  VR@FAR=0.1\%(\%) & Dimension \\
\hline
%PCA+Sym+PCA\cite{DBLP:conf/cvpr/LiYLL13} &23.70\%$\pm$1.89\%  & 19.27\% &-\\
%LCFS\cite{DBLP:journals/tifs/JinLR15}  &35.40\%$\pm$2.80\%  & 16.74\%&-\\
%C-DFD\cite{DBLP:journals/tifs/JinLR15} &65.80\%$\pm$1.60\% & 46.30\%&-\\
%DSIFT+PCA+LDA \cite{DBLP:conf/icpr/DhamechaSSV14} &73.28\%$\pm$1.10\% & - & -\\
CDFL   &71.50$\pm$1.40 &55.10 &1000\\
Gabor+Remove 20 PCs   &75.54$\pm$0.75 &71.40$\pm$1.21 &-\\
Gabor+RBM+Remove 11PCs  &86.16$\pm$0.98 &81.29$\pm$1.82 &$80\times176=14080$\\
NIR-VIS reconstruction+UDP   &78.46$\pm$1.67 &85.80 &$32\times32=1024$\\
HFR-CNN & 85.90$\pm$0.90 & 78.00 & 320 \\
TRIVET & 95.74$\pm$0.52 & 91.03$\pm$1.26 & 256 \\
IDR-128 & 97.33$\pm$0.43 & 95.73$\pm$0.73 & 128 \\
\hline
VGG  & 62.09$\pm$1.88& 39.72$\pm$2.85 & 4096\\
SeetaFace  & 68.03$\pm$1.66& 58.75$\pm$2.26 & 2048\\
CenterLoss  &87.69$\pm$1.45 & 69.72$\pm$2.07 & $2\times 512=1024$\\
Light CNN & 91.88$\pm$0.58& 85.31$\pm$0.95 & 256 \\
\hline
CDL &\textbf{98.62$\pm$0.20} & \textbf{98.32$\pm$0.05} & 256\\
\hline
\end{tabular}
\label{tab:state_of_the_art}

\end{table*}
%
%\begin{table}[tbp]
%\centering
%\caption{The comparison of the performance initialized by different pre-trained models. The items of CASIA-WebFace and MS-Celeb-1M refers to the basic model performance and }
%\begin{tabular}{|c|c|c|}
%\hline
%Method & Rank-1 & VR@FAR=0.1\%  \\
%\hline
%CASIA-WebFace & 85.45\%$\pm$ 1.45\%& 70.97\%$\pm$1.85\%\\
%Fine-tune  & 92.02\%$\pm$ 0.72\%& 88.67\%$\pm$1.26\%\\
%\hline
%MS-Celeb-1M &91.88\%$\pm$0.58\%  & 85.31\%$\pm$0.95\% \\
%Fine-tune & 97.15\%$\pm$0.66\%& 93.66\%$\pm$1.74\%\\
%\hline
%\end{tabular}
%\label{tab:compare_dataset}
%\end{table}

In Table~\ref{tab:compare}, the performance measures are reported for comparison, including verification rate(VR)@false acceptance rate(FAR). Firstly, comparing with performance of softmax and relevance constraint supervisory, the latter improves VR@FAR=0.1\% and VR@FAR=0.01\% by 0.91\% and 2.20\%, respectively.  It suggests that the usage of the relevance constraints are effective. And then, the performance of cross modal ranking is lower than that of softmax on VR@FAR=0.1\%(93.23\% vs 93.66\%) and VR@FAR=0.01\%(78.93\% vs 80.76\%). These results show that only using the cross modal ranking signal can not ensure large inter-class variations. As mentioned above, we observe that the rank-1 accuracy, VR@FAR=0.1\% and VR@FAR=0.01\% of CDL have been better than others. %The results suggest the effectiveness of the proposed CDL framework.

%Besides, we implement generic cross modal algorithms such as Maximum Mean Discrepancy (MMD) \cite{MMD_gretton2012kernel} and Central Moment Discrepancy (CMD) \cite{DBLP:journals/corr/ZellingerGLNS17} shown in Table \ref{tab:compare}. It is obvious that CDL outperforms MMD and CMD, since MMD and CMD only consider the differences between modalities rather than the properties of each identities, which are also important for heterogeneous face recognition.

%Besides, we further discuss the influence of the convolution parameters whether they are shared or not in CDL. If the parameters of convolution layers are not shared, it means that there are two feature extractors for NIR and VIS images, respectively. As is shown in Table~\ref{tab:compare}, we find the model which shares the convolution layers between different modalities obtains better performance than non-shared one (98.62\% vs 98.36\% for rank-1 and 98.32\% vs 98.02\% for VR@FAR=0.1\%). Since the shared convolutions between different modalities in CDL can reduce the number of parameters, they potentially alleviate the overfitting on a small number of training data.

%We also explore how the basic network affects the performance significantly. We implement two lightened CNN on CASIA-WebFace \cite{yi2014learning} and MS-Celeb-1M \cite{DBLP:journals/corr/GuoZHHG16} VIS datasets. The CASIA-WebFace database contains 10,575 identities totally about 500K images, while there are 8.5M face images about 100K identities in MS-Celeb-1M.
%
\begin{figure}
\centering
\includegraphics[width=0.35\textwidth]{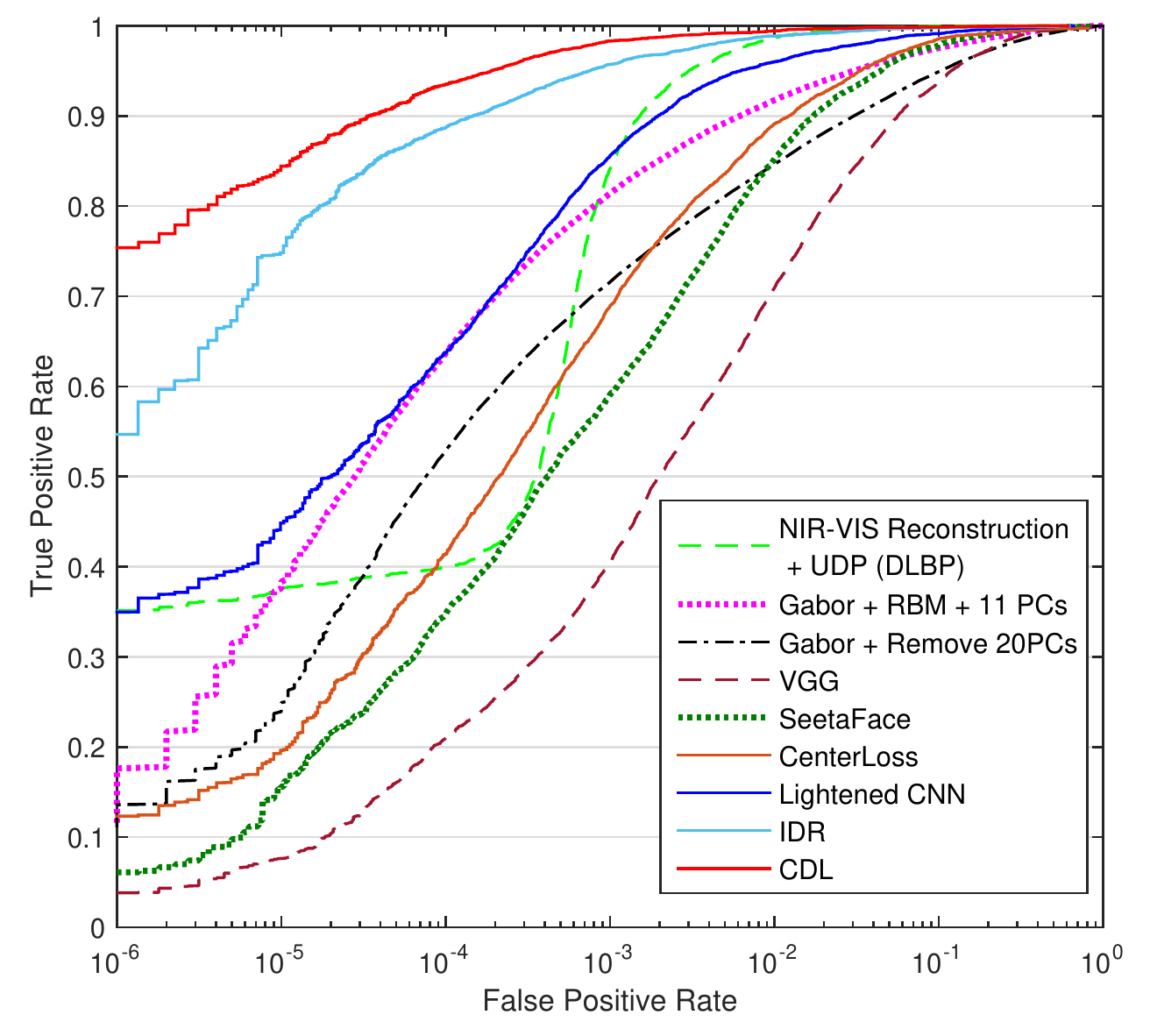}
\caption{ROC curves of different methods on the CASIA NIR-VIS 2.0 database.}
\label{fig:roc}
\end{figure}

\subsection{Method Comparison}

We compare the performance of CDL with other state-of-the-art NIR-VIS face recognition methods, including coupled discriminant feature learning (CDFL)\cite{DBLP:journals/tifs/JinLR15}, NIR-VIS reconstruction + UDP\cite{DBLP:conf/cvpr/Juefei-XuPS15} and Gabor + RBM + Remove 11PCs\cite{DBLP:conf/fgr/YiLL15} and six CNN-based methods such as VGG\cite{parkhi2015deep}, SeetaFace\cite{liu2016viplfacenet}, CenterLoss\cite{wen2016discriminative}, Light CNN\cite{wulight}, HFR-CNN\cite{DBLP:conf/eccv/SaxenaV16}, TRIVET\cite{DBLP:conf/icb/LiuSWT16} and IDR\cite{DBLP:conf/aaai/RanIDR17}.

Fig.~\ref{fig:roc} plots the receiver operating characteristic (ROC) curves of our CDL and its eight top competitors. For a better illustration, especially for a low FAR interval, we do not report other ROC curves if these curves are low and use semi-logarithmic coordinate to show the curves. Obviously, our CDL consistently outperforms Gabor+Remove 20 PCs\cite{DBLP:conf/fgr/YiLL15}, Gabor+RBM+Remove 11PCs\cite{DBLP:conf/fgr/YiLL15}, NIR-VIS reconstruction+UDP\cite{DBLP:conf/cvpr/Juefei-XuPS15}, IDR \cite{DBLP:conf/aaai/RanIDR17} and other general face recognition methods when FAR is lower than 1\%. The CDL obtains around 75\% at FAR=0.0001\% while IDR \cite{DBLP:conf/aaai/RanIDR17} gets only 55\% and others are even lower than 40\%. These results further show that CDL can obtain the discriminative feature representation and correctly classify some difficult NIR-VIS pairs.

\begin{table}[tbp]
\centering
\caption{The comparison of rank-1 and VR@FAR=1\% for the IIIT-D Sketch Database.}
\begin{tabular}{|c|c|c|}
\hline
Method & Rank-1(\%) & FAR=1\%(\%)\\
\hline
Original WLD & 74.34 & - \\
SIFT & 76.28 & -\\
EUCLBP & 79.36 & -\\
LFDA & 81.43 & -\\
MCWLD & 84.24& -\\
\hline
VGG  & 80.89 & 72.08\\
CenterLoss  & 84.07 & 76.20\\
Light CNN & 84.07 & 75.30\\
CDL & \textbf{85.35} & \textbf{82.52}\\
\hline
\end{tabular}
\label{tab:sketch}
\vspace{-4mm}
\end{table}

%\begin{figure}
%\centering
%\includegraphics[width=0.35\textwidth]{sketch_ROC.pdf}
%\caption{ROC curves of different CNN-based methods on the IIIT-D Sketch Database.}
%\label{fig:roc_sketch}
%\vspace{-4mm}
%\end{figure}

Table~\ref{tab:state_of_the_art} shows the rank-1 accuracy and VR@FAR=0.1\% of state-of-the-art methods. Our CDL improves the best rank-1 accuracy from 97.33\% to 98.62\% compared with IDR. Besides, VR@FAR=0.1\% has also significantly improved from 95.73\% to 98.32\%, suggesting that CDL can obtain more discriminative features than its competitors. All of these results demonstrate that the proposed CDL is effective for heterogeneous face recognition, and can learn a compact and modality invariant feature representation.

We also compare CDL with other open source CNN face recognition solutions including VGG \cite{parkhi2015deep}, SeetaFace \cite{liu2016viplfacenet}, Light CNN \cite{wulight} and CenterLoss \cite{wen2016discriminative}. CDL archives the highest performance on rank-1 accuracy (98.62\% vs 91.88\%) and VR@FAR=0.1\% (98.32\% vs 85.31\%) as shown in Table~\ref{tab:state_of_the_art}. Obviously, simply applying CNN to heterogeneous face recognition is not effective.

We also evaluate CDL on the viewed sketch-photo face recognition database. As is shown in Table \ref{tab:sketch}, CDL obtains 85.35\% rank-1 accuracy which outperforms other viewed sketch-photo methods on the IIIT-D Sketch Database. Note that we employ CUFSF as the training dataset on different CNN-based methods. Table \ref{tab:sketch} shows the performance of VGG, CenterLoss, Lightened CNN and CDL on the IIIT-D Sketch Database. It is obvious that the proposed CDL is better than other CNN-based methods on a small number of training samples. Besides, if there are more sketch-photo images (not only two images in each identities), the performance of CDL will be further improved.

\section{Conclusions}

%In this paper, we proposed a coupled deep learning (CDL) framework for heterogeneous face recognition by introducing low-rank relevance constraint and cross modal ranking into CNN. The CDL is initialized by pre-trained VIS face recognition model and the low rank constraint is introduced to increase the correlation between two modalities as the supervised signal. An alternating iterative optimization method is utilized for back-propagation to optimize the CNN. Moreover, the cross modal ranking is further improved the discriminant of the CNN representation. Experimental results on the challenging CASIA NIR-VIS 2.0 face recognition database show that our CDL outperforms other state-of-the-art methods.
In this paper, we propose a coupled deep learning (CDL) framework for heterogeneous face recognition by introducing trace norm and block-diagonal relevance constraints, and cross modal ranking into CNN. CDL is an effective way for the limited number of training samples. The low rank and block-diagonal constraints are utilized to increase the correlation between two modalities as the supervised signal and the cross modal ranking has been used to further improve the discrimination of CDL. Moreover, an alternating iterative optimization method has been developed for back-propagation to optimize an end-to-end CNN. Experimental results on the challenging CASIA NIR-VIS 2.0 face recognition database and viewed sketch-photo databases show that our CDL significantly outperforms other state-of-the-art heterogeneous face recognition methods.

\section*{Acknowledgment}

This work is partially funded by National Natural Science Foundation of China (Grant No.61473289, 61622310) and the State Key Development Program (Grant No. 2016YFB1001001).

%\begin{table*}[htbp]
%\centering
%\caption{The comparison of Rank-1 accuracy ($\pm$ standard variation) and VR@FAR=0.1\% ($\pm$ standard variation) on CASIA NIR-VIS 2.0 database.}
%\begin{tabular}{|c|c|c|c|}
%\hline
%Method  & Rank-1 accuracy &  VR@FAR=0.1\%  \\
%\hline
%\hline
%CDFL  &71.50\% &55.10\% \\
%Gabor+Remove 20 PCs  &75.54\% &71.40\% \\
%Gabor+RBM+Remove 11PCs  &86.16\% &81.29\% \\
%NIR-VIS reconstruction+UDP  &78.46\%\% &85.80\% \\
%\hline
%\hline
%VGG & 62.09\%& 39.72\% \\
%SeetaFace & 68.03\%& 58.75\% \\
%Lightened CNN & 91.88\%& 85.31\% \\
%CenterLoss (ECCV 16) &87.69\% & 69.72\% \\
%\hline
%\hline
%%Basic model &91.88\%$\pm$0.58\%  & 85.31\%$\pm$0.95\%& 256\\
%CL-CNN &\textbf{98.62\%} & \textbf{98.32\%} \\
%\hline
%\end{tabular}
%\label{tab:state_of_the_art}
%\end{table*}

{
\bibliographystyle{aaai}
\bibliography{egbib}

\begin{thebibliography}{}

\bibitem[\protect\citeauthoryear{Bhatt \bgroup et al\mbox.\egroup
  }{2012}]{bhatt2012memetic}
Bhatt, H.~S.; Bharadwaj, S.; Singh, R.; and Vatsa, M.
\newblock 2012.
\newblock Memetic approach for matching sketches with digital face images.
\newblock Technical report, IIITD-TR-2011-006.

\bibitem[\protect\citeauthoryear{Dhamecha \bgroup et al\mbox.\egroup
  }{2014}]{dhamecha2014effectiveness}
Dhamecha, T.~I.; Sharma, P.; Singh, R.; and Vatsa, M.
\newblock 2014.
\newblock On effectiveness of histogram of oriented gradient features for
  visible to near infrared face matching.
\newblock In {\em ICPR}.

\bibitem[\protect\citeauthoryear{Goswami \bgroup et al\mbox.\egroup
  }{2011}]{goswami2011evaluation}
Goswami, D.; Chan, C.~H.; Windridge, D.; and Kittler, J.
\newblock 2011.
\newblock Evaluation of face recognition system in heterogeneous environments
  (visible vs nir).
\newblock In {\em ICCV Workshop}.

\bibitem[\protect\citeauthoryear{Grave, Obozinski, and
  Bach}{2011}]{DBLP:conf/nips/GraveOB11}
Grave, E.; Obozinski, G.; and Bach, F.~R.
\newblock 2011.
\newblock Trace lasso: a trace norm regularization for correlated designs.
\newblock In {\em NIPS}.

\bibitem[\protect\citeauthoryear{Gretton \bgroup et al\mbox.\egroup
  }{2012}]{MMD_gretton2012kernel}
Gretton, A.; Borgwardt, K.~M.; Rasch, M.~J.; Sch{\"o}lkopf, B.; and Smola, A.
\newblock 2012.
\newblock A kernel two-sample test.
\newblock {\em JMLR} 13(3):723--773.

\bibitem[\protect\citeauthoryear{Guo \bgroup et al\mbox.\egroup
  }{2016}]{DBLP:journals/corr/GuoZHHG16}
Guo, Y.; Zhang, L.; Hu, Y.; He, X.; and Gao, J.
\newblock 2016.
\newblock Ms-celeb-1m: {A} dataset and benchmark for large-scale face
  recognition.
\newblock In {\em ECCV}.

\bibitem[\protect\citeauthoryear{He \bgroup et al\mbox.\egroup
  }{2017}]{DBLP:conf/aaai/RanIDR17}
He, R.; Wu, X.; Sun, Z.; and Tan, T.
\newblock 2017.
\newblock Learning invariant deep representation for nir-vis face recognition.
\newblock In {\em AAAI}.

\bibitem[\protect\citeauthoryear{Hou, Yang, and Wang}{2014}]{hou2014domain}
Hou, C.-A.; Yang, M.-C.; and Wang, Y.-C.~F.
\newblock 2014.
\newblock Domain adaptive self-taught learning for heterogeneous face
  recognition.
\newblock In {\em ICPR}.

\bibitem[\protect\citeauthoryear{Jin, Lu, and
  Ruan}{2015}]{DBLP:journals/tifs/JinLR15}
Jin, Y.; Lu, J.; and Ruan, Q.
\newblock 2015.
\newblock Coupled discriminative feature learning for heterogeneous face
  recognition.
\newblock {\em IEEE TIFS} 10(3):640--652.

\bibitem[\protect\citeauthoryear{Juefei{-}Xu, Pal, and
  Savvides}{2015}]{DBLP:conf/cvpr/Juefei-XuPS15}
Juefei{-}Xu, F.; Pal, D.~K.; and Savvides, M.
\newblock 2015.
\newblock {NIR-VIS} heterogeneous face recognition via cross-spectral joint
  dictionary learning and reconstruction.
\newblock In {\em CVPR Workshop}.

\bibitem[\protect\citeauthoryear{Kan, Shan, and
  Chen}{2016}]{DBLP:conf/cvpr/KanSC16}
Kan, M.; Shan, S.; and Chen, X.
\newblock 2016.
\newblock Multi-view deep network for cross-view classification.
\newblock In {\em CVPR}.

\bibitem[\protect\citeauthoryear{Klare and Jain}{2010}]{klare2010heterogeneous}
Klare, B., and Jain, A.~K.
\newblock 2010.
\newblock Heterogeneous face recognition: Matching nir to visible light images.
\newblock In {\em ICPR}.

\bibitem[\protect\citeauthoryear{Lei \bgroup et al\mbox.\egroup
  }{2008}]{lei2008CCA_mapping}
Lei, Z.; Bai, Q.; He, R.; and Li, S.~Z.
\newblock 2008.
\newblock Face shape recovery from a single image using cca mapping between
  tensor spaces.
\newblock In {\em CVPR}.

\bibitem[\protect\citeauthoryear{Li \bgroup et al\mbox.\egroup
  }{2013}]{DBLP:conf/cvpr/LiYLL13}
Li, S.~Z.; Yi, D.; Lei, Z.; and Liao, S.
\newblock 2013.
\newblock The {CASIA} {NIR-VIS} 2.0 face database.
\newblock In {\em CVPR Workshop}.

\bibitem[\protect\citeauthoryear{Liao \bgroup et al\mbox.\egroup
  }{2009}]{liao2009heterogeneous}
Liao, S.; Yi, D.; Lei, Z.; Qin, R.; and Li, S.~Z.
\newblock 2009.
\newblock Heterogeneous face recognition from local structures of normalized
  appearance.
\newblock In {\em ICB}.

\bibitem[\protect\citeauthoryear{Lin and Tang}{2006}]{lin_tang2006}
Lin, D., and Tang, X.
\newblock 2006.
\newblock Inter-modality face recognition.
\newblock In {\em ECCV}.

\bibitem[\protect\citeauthoryear{Liu \bgroup et al\mbox.\egroup
  }{2005}]{liu2005nonlinear}
Liu, Q.; Tang, X.; Jin, H.; Lu, H.; and Ma, S.
\newblock 2005.
\newblock A nonlinear approach for face sketch synthesis and recognition.
\newblock In {\em CVPR}.

\bibitem[\protect\citeauthoryear{Liu \bgroup et al\mbox.\egroup
  }{2016a}]{DBLP:conf/icb/LiuSWT16}
Liu, X.; Song, L.; Wu, X.; and Tan, T.
\newblock 2016a.
\newblock Transferring deep representation for {NIR-VIS} heterogeneous face
  recognition.
\newblock In {\em International Conference on Biometrics}.

\bibitem[\protect\citeauthoryear{Liu \bgroup et al\mbox.\egroup
  }{2016b}]{liu2016viplfacenet}
Liu, X.; Kan, M.; Wu, W.; Shan, S.; and Chen, X.
\newblock 2016b.
\newblock {VIPLFaceNet}: An open source deep face recognition sdk.
\newblock {\em Frontiers of Computer Science}.

\bibitem[\protect\citeauthoryear{Ngiam \bgroup et al\mbox.\egroup
  }{2011}]{DBLP:conf/icml/NgiamKKNLN11}
Ngiam, J.; Khosla, A.; Kim, M.; Nam, J.; Lee, H.; and Ng, A.~Y.
\newblock 2011.
\newblock Multimodal deep learning.
\newblock In {\em ICML}.

\bibitem[\protect\citeauthoryear{Parkhi, Vedaldi, and
  Zisserman}{2015}]{parkhi2015deep}
Parkhi, O.~M.; Vedaldi, A.; and Zisserman, A.
\newblock 2015.
\newblock Deep face recognition.
\newblock {\em BMVC}.

\bibitem[\protect\citeauthoryear{Saxena and
  Verbeek}{2016}]{DBLP:conf/eccv/SaxenaV16}
Saxena, S., and Verbeek, J.
\newblock 2016.
\newblock Heterogeneous face recognition with cnns.
\newblock In {\em ECCV Workshop}.

\bibitem[\protect\citeauthoryear{Schroff, Kalenichenko, and
  Philbin}{2015}]{schroff2015facenet}
Schroff, F.; Kalenichenko, D.; and Philbin, J.
\newblock 2015.
\newblock Facenet: {A} unified embedding for face recognition and clustering.
\newblock In {\em CVPR}.

\bibitem[\protect\citeauthoryear{Socolinsky and
  Selinger}{2002}]{socolinsky2002TIR_analysis}
Socolinsky, D.~A., and Selinger, A.
\newblock 2002.
\newblock A comparative analysis of face recognition performance with visible
  and thermal infrared imagery.
\newblock Technical report, DTIC Document.

\bibitem[\protect\citeauthoryear{Wang and
  Tang}{2009}]{DBLP:journals/pami/WangT09}
Wang, X., and Tang, X.
\newblock 2009.
\newblock Face photo-sketch synthesis and recognition.
\newblock {\em IEEE TPAMI} 31(11):1955--1967.

\bibitem[\protect\citeauthoryear{Wang \bgroup et al\mbox.\egroup
  }{2013}]{DBLP:conf/iccv/WangHWWT13}
Wang, K.; He, R.; Wang, W.; Wang, L.; and Tan, T.
\newblock 2013.
\newblock Learning coupled feature spaces for cross-modal matching.
\newblock In {\em ICCV}.

\bibitem[\protect\citeauthoryear{Wang \bgroup et al\mbox.\egroup
  }{2015}]{DBLP:conf/icml/WangALB15}
Wang, W.; Arora, R.; Livescu, K.; and Bilmes, J.~A.
\newblock 2015.
\newblock On deep multi-view representation learning.
\newblock In {\em ICML}.

\bibitem[\protect\citeauthoryear{Wen \bgroup et al\mbox.\egroup
  }{2016}]{wen2016discriminative}
Wen, Y.; Zhang, K.; Li, Z.; and Qiao, Y.
\newblock 2016.
\newblock A discriminative feature learning approach for deep face recognition.
\newblock In {\em ECCV}.

\bibitem[\protect\citeauthoryear{Wu \bgroup et al\mbox.\egroup
  }{2016}]{wulight}
Wu, X.; He, R.; Sun, Z.; and Tan, T.
\newblock 2016.
\newblock A light cnn for deep face representation with noisy labels.
\newblock {\em arXiv preprint arXiv:1511.02683}.

\bibitem[\protect\citeauthoryear{Yan and
  Mikolajczyk}{2015}]{DBLP:conf/cvpr/YanM15}
Yan, F., and Mikolajczyk, K.
\newblock 2015.
\newblock Deep correlation for matching images and text.
\newblock In {\em CVPR}.

\bibitem[\protect\citeauthoryear{Yi \bgroup et al\mbox.\egroup
  }{2007}]{yi2007face_NIS-VIR}
Yi, D.; Liu, R.; Chu, R.; Lei, Z.; and Li, S.~Z.
\newblock 2007.
\newblock Face matching between near infrared and visible light images.
\newblock In {\em ICB}.

\bibitem[\protect\citeauthoryear{Yi, Lei, and Li}{2015}]{DBLP:conf/fgr/YiLL15}
Yi, D.; Lei, Z.; and Li, S.~Z.
\newblock 2015.
\newblock Shared representation learning for heterogenous face recognition.
\newblock In {\em FG}.

\bibitem[\protect\citeauthoryear{Zellinger \bgroup et al\mbox.\egroup
  }{2017}]{DBLP:journals/corr/ZellingerGLNS17}
Zellinger, W.; Grubinger, T.; Lughofer, E.; Natschl{\"{a}}ger, T.; and
  Saminger{-}Platz, S.
\newblock 2017.
\newblock Central moment discrepancy {(CMD)} for domain-invariant
  representation learning.
\newblock In {\em ICLR Workshop}.

\bibitem[\protect\citeauthoryear{Zhang, Wang, and
  Tang}{2011}]{DBLP:conf/cvpr/ZhangWT11}
Zhang, W.; Wang, X.; and Tang, X.
\newblock 2011.
\newblock Coupled information-theoretic encoding for face photo-sketch
  recognition.
\newblock In {\em CVPR}.

\bibitem[\protect\citeauthoryear{Zhu \bgroup et al\mbox.\egroup
  }{2014}]{zhu2014matching}
Zhu, J.-Y.; Zheng, W.-S.; Lai, J.-H.; and Li, S.~Z.
\newblock 2014.
\newblock Matching nir face to vis face using transduction.
\newblock {\em IEEE TIFS} 9(3):501--514.

\end{thebibliography}
}

\end{document}